\documentclass[runningheads]{llncs}

\usepackage{graphicx}
\usepackage{amsmath}
\usepackage{amsfonts}
\usepackage{xcolor}
\usepackage{caption}
\usepackage{subcaption}
\usepackage{url}

\newcommand\defeq{\mathrel{\stackrel{\makebox[0pt]{\mbox{\normalfont\scriptsize def}}}{:=}}}
\newcommand{\sig}[1]{{\textsf{{#1}}}}
\newcommand{\ct}{\textsf{ct}}

\newcommand{\Comment}[1]{}

\begin{document}
\title{Safety Performance of Neural Networks in the Presence of Covariate Shift}
\author{
Chih-Hong~Cheng\inst{1}
\and
Harald Ruess\inst{2}
\and
Konstantinos Theodorou\inst{1}
}

\authorrunning{Cheng, Ruess, Theodorou}

\institute{Fraunhofer IKS, Munich, Germany \\
\and
fortiss, Munich, Germany \\
\email{\{chih-hong.cheng,konstantinos.theodorou\}@iks.fraunhofer.de, ruess@fortiss.org}}

\maketitle  

\begin{abstract}
Covariate shift may impact the operational safety performance of neural networks. A re-evaluation of the safety performance, however, requires collecting new operational data and creating corresponding ground truth labels, which often is not possible during operation. We are therefore proposing to reshape the initial test set, as used for the safety performance evaluation prior to deployment, based on an approximation of the operational data. This approximation is obtained by observing and learning the distribution of activation patterns of neurons in the network during operation. The reshaped test set reflects the distribution of neuron activation values as observed during operation, and may therefore be used for re-evaluating safety performance in the presence of covariate shift. First, we derive conservative bounds on the values of neurons by applying finite binning and static dataflow analysis. Second, we formulate a {\em mixed integer linear programming} (MILP) constraint for constructing the minimum set of data points to be removed in the test set, such that the difference  between the discretized test and operational distributions is bounded. We discuss potential benefits and limitations of this constraint-based approach based on our initial experience with an implemented research prototype.
\keywords{distribution reshaping \and machine learning \and  MILP \and performance estimation}
\end{abstract}

\section{Introduction}
\label{sec:introduction}

{\em Covariate shift} in machine-learned systems occurs when the input distribution changes between training and operation stages~\cite{quinonero2008dataset}\@.
This phenomenon is present in most applications of machine learning, as training sets usually do not sufficiently reflect the complexity of real-world operational contexts and their potential changes.
This kind of dataset shift is also a major concern in transfer learning when exposing machine-learned systems to solving different tasks~\cite{kouw2019introduction}\@.

There is a fundamental dichotomy between covariate shift of machine-learned systems during operation and their underlying safety requirements, which require demonstrating given {\em safety performance indicators} (SPIs)~\cite{koopman2020positive,UL4600} prior to deployment.
The challenge we are tackling therefore is to incorporate the possibility of operational covariate shift into safety assurance arguments for safety-related neural networks. 
Hereby, we assume that the initial test data for SPI evaluation is known but the operational data is unknown, since collecting operational data and creating corresponding ground truth labels is usually not possible during operation.

In tackling this challenge, we develop a specialized online monitoring technique for estimating the 
change of values of SPIs due to covariate shift. 
For practical purposes, we restrict ourselves to 
feed-forward deep neural networks (DNN), and
we assume that the values of monitored neurons (in the feature space)
of this DNN adequately reflect the input data distribution. 
Now, for each monitored neuron, one constructs the histogram of distributions based on binning, 
and distribution shift is observed by comparing the shape of two such histograms. Such information abstracts the details of the observed input in operation, thereby bypassing practical limitations where arbitrarily initiating a data collection regime when the DNN is integrated into an application has technical and societal constraints. 
One of our measures of similarity, 
called $\kappa$-KL similarity, is inspired by the Kullback-Leibler divergence.
The purpose of a second measure, called $\epsilon$-portion similarity, 
is to characterize bounded differences.
Key to our approach is that SPI estimations are
reduced to constructing a subset of the test data that matches the similarity measure demonstrated by the operation, followed by the recomputation of the SPI on this subset. 
For $\epsilon$-portion similarity, when the input is of bounded range
we introduce a {\em mixed integer linear programming} (MILP) encoding with 0-1 variables, whereby an upper limit on the number of bins can be 
obtained from static dataflow analyis.

In Section~\ref{sec:related} we review and compare with most closely related work. Section~\ref{sec:algorithm} defines  $\epsilon$-portion similarity  for measuring the similarity of distributions based on neuron activations. Next, Section~\ref{sec:encoding} describes distribution reshaping for  $\epsilon$-portion similarity via test set reduction together its encoding in MILP. This algorithm is evaluated in Section~\ref{sec:evaluation} based on standard machine learning benchmarks, and we conclude in Section~\ref{sec:conclusion} with discussing the potential benefits and current limitations of our approach.

\section{Related Work}
\label{sec:related}

The divergence between a source and a target distribution, as obtained, say, by dataset shift is usually measured in terms of {\em mutual information} or KL divergence~\cite{quinonero2008dataset}\@. 
Unfortunately, measuring mutual information from finite
data is a notoriously difficult estimation problem~\cite{kraskov2004estimating,gao2015efficient}, 
and there are statistical limitations on measuring lower bounds on KL divergence from finite data~\cite{mcallester2020formal}\@. 
Since our techniques are intended to be applied during operation, we are therefore measuring covariate shift only indirectly by observing and comparing abstracted distributions on corresponding neuronal excitements. 
This kind of indirect measurement does not require the original detailed operational data (e.g., images) to be available.

Covariate shift in machine-learned systems may be corrected using, say, {\em weighted empirical risk minimization}~\cite{vogel2020weighted}, which is based on retraining the system with a calibrated loss function based on the ratio of source and the target distribution of inputs.
Retraining of safety-related machine-learned components, however, is problematic, 
and the best we can do is to adequately measure the potential drop of relevant safety performance indicators which are due to covariate shift.
This implies that when data in operation is only made available as abstracted distributions, we need to reshape the test data to create an estimation.  
$\epsilon$-portion similarity, as developed here, 
is to practically consider the non-linearity 
caused by KL-divergence as the similarity measure, thereby enabling a reduction to MILP\@.

Within the research field of safe autonomous 
driving, {\em leading measures}~\cite{fraade2018measuring} are proactive indicators that assess prevention efforts and can be observed and evaluated before a crash occurs. Our approach to SPI re-estimation under distribution reshaping yields a leading measure  on the level of a machine-learned component. Our developments also go beyond {\em out-of-distribution} techniques for detecting outliers with respect to training inputs~\cite{lust2020survey,ruff2021unifying}\@,
 as we are constructing an aggregated SPI against all observed data, where even when every data point being observed in operation is {\em within-distribution}, this does not imply that the SPI will be the same.

\section{Distribution Similarity based on Neuron Values}
\label{sec:algorithm}

$\mathcal{D}_{op}$ denotes the multiset of data points, which are collected during operation, and~$\mathcal{D}_{test}$ is the multiset of data points used in the (safety) performance evaluation. 
We assume as given a feed-forward {\em deep neural network} (DNN)
$F \defeq f^L \circ  \ldots \circ f^1$, which is 
composed of layers~$1$ through $L$\@. 
Each {\em layer} $l_i$ is a function $\mathbb{R}^{d_{i-1}} \to \mathbb{R}^{d_{i}}$,
with $d_i$ the dimensions of vectors.
Layers consists of a set of neurons
for computing a weighted linear sum from the input of the previous layer, followed by applying some
monotonically non-decreasing activation function, 
such as {\em ReLU}, {\em Leaky ReLU}, and {\em tanh}\@.
Without loss of generality, the neurons in $F$ are fully connected with subsequent layers.
The notation $l_A \in \{1, \ldots, L\}$ indicates 
the chosen layer for analyzing distribution similarity between $\mathcal{D}_{op}$ and $\mathcal{D}_{test}$\@. 
For a data point $\sig{in}$, the output at the $l$-th layer is 
the vector $F^l(\sig{in}) := f^l (f^{l-1}(\ldots f^1(\sig{in})))$
of dimension $d_l$\@. 
Now, $F^l_i(\sig{in})$ projects the $i$-th output from $F^l(\sig{in})$,
and $f^l_i$ is the output of the $i$-th neuron at the $l$-th layer; that is, given an input $\sig{in}$, $f^l_i$ takes $F^{l-1}(\sig{in})$ as input and produces $f^l_i(F^{l-1}(\sig{in}))$ which equals $F^l_i(\sig{in})$\@. 
Finally, all inputs 
are bounded by an interval $[v_{min}, v_{max}]$, where
$v_{min}$, $v_{max}$ are fixed constants. 
In other words,  $\sig{in} \in [v_{min}, v_{max}]^{d_0}$ and $\mathcal{D}_{op}, \mathcal{D}_{test} \subseteq [v_{min}, v_{max}]^{d_0}$\@. 
For a multiset $\mathcal{D}$ of data points and given DNN $F$, we define 
   \begin{align}
V^l_i(\mathcal{D}) &\defeq \langle F^l_i(\sig{in}) \;|\; \sig{in} \in \mathcal{D} \rangle
    \end{align}
to be the multiset of all values of the $i$-th neuron value at layer $l$ for all inputs in $\mathcal{D}$.

\begin{definition}\label{def:binning}
For a natural number $N > 0$, a positive real number $\Delta$ and a real number $c \in \mathbb{R}$, the \textbf{$(c,\Delta,N)$-binning function} $b^{c,\Delta}_N: [c, c + (N+1)\Delta] \rightarrow \{0, 1,  \ldots, N \}$ 
is defined as follows:

\vspace{-5mm}
\begin{equation}
b^{c,\Delta}_N(x) =
\begin{cases}
0 & \text{if $x \in [c, c + \Delta]$} \\
$j$ & \text{else if $x \in (c + j \Delta, c + (j+1)\Delta]$, for any $j \in \{1, \ldots, N\}$} 
\end{cases}
\end{equation}
\end{definition}

We apply each element in $V^{l_A}_i(\mathcal{D}_{op})$ with a binning function in order to derive another multiset 
\begin{align}
B^{l_A}_i(F, \mathcal{D}_{op}) \defeq \langle b^{c,\Delta}_N(F^{l_A}_i(\sig{in}))\; |\; \sig{in} \in \mathcal{D}_{op} \rangle\mbox{\@.}
\end{align}
This requires, however, that for all $\sig{in}\in \mathcal{D}_{op}$,  $F^{l_A}_i(\sig{in}) \in [c, c +  (N+1)\Delta]$. 
Provided that any input is bounded where $\sig{in} \in [v_{min}, v_{max}]^{d_0}$, we have the following property.    

\begin{lemma}\label{lemma:dataflow}
Let $\Delta$ be a positive constant. Provided that $\mathcal{D}_{op}, \mathcal{D}_{test} \subseteq [v_{min}, v_{max}]^{d_0}$ and $F$ is implemented layer-wise with each neuron $f^l_i$ implemented by (1) performing a weighted linear sum from the previous layer, followed by (2) applying monotonically non-decreasing computational activation function,
there exists a constant $c\in \mathbb{R}$ and $N \in \mathbb{N}$ such that for all $i \in \{1, \ldots, d_l\}$, $F^{l_A}_i(\sig{in})  \in [c, c +  N\Delta]$, where $c$ and $N$ can be computed in time linear to the number of neurons. 
\end{lemma}

\proof (Sketch) This is based on the known result in neural network verification using abstract interpretation~\cite{gehr2018ai2,cheng2017maximum,gowal2019scalable}, where provided that input is bounded, and $F$ is implemented with (1) and (2), one can apply computationally efficient interval-bound propagation (boxed abstraction)~\cite{cheng2017maximum,gowal2019scalable} to derive a conservative minimum and maximum bound $[v_{i,min}, v_{i, max}]$ such that $\{ F^{l_A}_i(\sig{in}) \;|\; \sig{in}\in [v_{min}, v_{max}]^{d_0} \} \subseteq [v_{i,min}, v_{i, max}]$, where the interval-bound analysis is done in time linear to the total amount of neurons (see Example~\ref{example.propagation} for illustration).\footnote{This is because intuitively, a deep neural network is nothing different than a program without loop. Therefore, interval-bound propagation only requires a single pass.} With the following value assignment for~$c$ and~$N$, the lemma then holds. 

\vspace{-5mm}
\begin{align}
    c := \min (v_{1, min}, \ldots, v_{d_l, min}) \\
    N := \lceil \frac{\max (v_{1, max}, \ldots, v_{d_l, max}) - c}{\Delta}\rceil 
\end{align}

\vspace{-2mm}
\qed

\begin{figure}[t]

    \centering
\vspace{-2mm}
    \includegraphics[width=0.5\textwidth]{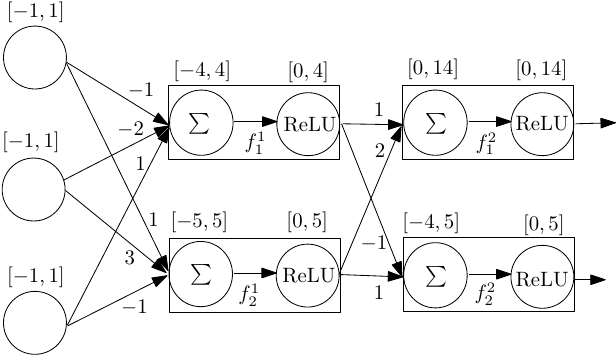}
    \caption{An example of using bound propagation to conservatively estimate~$c$ and~$N$.}
    \label{fig:bound.propagation}
    \vspace{-5mm}
\end{figure}

\vspace{-2mm}
\begin{example}

Consider the example in Figure~\ref{fig:bound.propagation}, where we wish to perform the analysis at layer~$l_A = 2$. Assume $\Delta = 3$, and for each input at the left, it has the domain $[-1, 1]$, i.e., $\mathcal{D}_{op}, \mathcal{D}_{test} \subseteq [-1, 1]^3$. For each neuron ($f^1_1$, $f^1_2$, $f^2_1$, $f^2_2$), its computation is completed by first performing a weighted sum (the corresponding weight is attached in the edge) followed by the nonlinear activation ReLU ($ReLU(x) \defeq \max(0, x)$). The result of interval-bound propagation provides us a conservative estimate $v_{1, min} = v_{2, min} =0$, $v_{1, max}=14$, $v_{2, max}=5$. Therefore, $c=\min (0, 0) = 0$ and $N = \lceil  \frac{\max(14, 5)-0}{3}\rceil = 5$. 
\label{example.propagation} 
\end{example}

By applying each element in $V^{l_A}_i(\mathcal{D}_{op})$ with the binning function created using Lemma~\ref{lemma:dataflow}, one derives another multiset $B^{l_A}_i(F, \mathcal{D}_{op}) \defeq \langle b^{c,\Delta}_N(F^{l_A}_i(\sig{in}))\; |\; \sig{in} \in \mathcal{D}_{op} \rangle$. Analogously,  define $B^{l_A}_i(F, \mathcal{D}_{test})$ to abbreviate $ \langle b^{c,\Delta}_N(F^{l_A}_i(\sig{in}))\; |\; \sig{in} \in \mathcal{D}_{test} \rangle$. Let $\ct(j, \mathcal{D})$ be the function that counts the number of elements in multiset~$\mathcal{D}$ having value~$j$, and $|\mathcal{D}|$ returns the size of the multiset. We now define two types of distribution similarity.

\begin{definition}\label{def:KL similar}
Given a positive constant $\kappa$, define $\mathcal{D}_{op}$ and $\mathcal{D}_{test}$ to be \textbf{$\kappa$-KL similar} (subject to DNN~$F$, layer index~$l_A$, and binning function~$b^{c, \Delta}_N$), if: 

\vspace{-5mm}

\begin{multline}
\forall i\in \{1, \ldots, d_l\}: \sum^N_{j=0}  \frac{\emph{\ct}(j, B^{l_A}_i(F, \mathcal{D}_{test}))}{|\mathcal{D}_{test}|} \ln\frac{\frac{\emph{\ct}(j, B^{l_A}_i(F, \mathcal{D}_{test}))}{|\mathcal{D}_{test}|} }{\frac{\emph{\ct}(j, B^{l_A}_i(F, \mathcal{D}_{op}))}{|\mathcal{D}_{op}|} }\leq \kappa 
\end{multline}
\end{definition}

\begin{definition}\label{def:portion.similar}
Given a positive constant $\epsilon$, define $\mathcal{D}_{op}$ and $\mathcal{D}_{test}$ to be  \textbf{$\epsilon$-portion similar} (subject to DNN~$F$, layer index~$l_A$, and binning function~$b^{c, \Delta}_N$) if:
\begin{multline}
\forall i\in \{1, \ldots, d_l\}, \forall j\in \{0, \ldots, N\}: \\
-\epsilon \leq \frac{\emph{\ct}(j, B^{l_A}_i(F, \mathcal{D}_{op}))}{|\mathcal{D}_{op}|} - \frac{\emph{\ct}(j, B^{l_A}_i(F, \mathcal{D}_{test}))}{|\mathcal{D}_{test}|} \leq \epsilon
\end{multline}
\end{definition}

\begin{table}[t]
    \centering
        \caption{Comparing two distribution similarity measures}
\begin{tabular}{ |c|c|c|c| } 
 \hline
Distribution similarity & Sensitive to $|\mathcal{D}_{op}|$ & Non-emptiness in bin & Dist. reshaping via MILP \\ \hline
$\kappa$-KL similar & no  & needed & no  \\ \hline
$\epsilon$-portion similar & no & not needed & yes  \\ \hline
\end{tabular}
    \label{table:summary.similarity}

    \vspace{-2mm}
\end{table}

Table~\ref{table:summary.similarity} summarizes the characteristics of our proposed distribution similarity measures.  The definition of $\kappa$-KL similarity is based on the well-known definition of KL-divergence in a discrete setting. As the computation $ \frac{\ct(j, B^{l_A}_i(F, \mathcal{D}_{test}))}{|\mathcal{D}_{test}|}$ and $ \frac{\ct(j, B^{l_A}_i(F, \mathcal{D}_{op}))}{|\mathcal{D}_{op}|}$ are only the relative frequencies in appearing to a particular bin, it is not sensitive to the size of $\mathcal{D}_{op}$. Nevertheless, due to the nonlinear logarithm function in the definition, the distribution shaping problem with minimum data point removal naturally cannot be formulated using MILP, which is in contrast to distribution reshaping using $\epsilon$-portion similarity. It is also crucial to notice that the standard KL divergence is defined when they have only non-zero entries; in our setup, this implies that all bins should be non-empty to ensure $\kappa$-KL similarity is well-defined. Here we omit details, but one may employ some heuristics: (1) Omit counting bin~$j$ where both $\mathcal{D}_{test}$ and $\mathcal{D}_{op}$ have no contribution to  bin~$j$. (2) When~$\mathcal{D}_{test}$ contributes to bin~$j$ but not $\mathcal{D}_{op}$, return ``undefined'' or ``$\infty$'' for the $\kappa$-KL similarity. 

Finally, when $\kappa$-KL similarity is well-defined, provided that $\epsilon$-portion-similarity holds, one can also derive a conservative~$\kappa$ value where the $\kappa$-KL-divergence-similarity measure is guaranteed to hold. It can be accomplished by considering a worst case where ``every'' bin in operation has the largest difference characterized by $\epsilon$-portion-similarity (which is overly conservative, as the sum of all bin-ratios for operation will not be~$1$), followed by feeding that information to the definition of $\kappa$-KL-divergence similarity, in order to derive a conservative bounding~$\kappa$ value. 

Note that for simplicity, we have prepared our formulation such that the distribution similarity is defined based on considering all neurons in layer~$l_A$ and having a unified binning function for all neurons. The constraint can be easily relaxed to allow distribution similarity to be considered only on a subset of neurons as well as neurons on different layers.

\begin{figure}[t]
    \centering
    \includegraphics[width=0.75\textwidth]{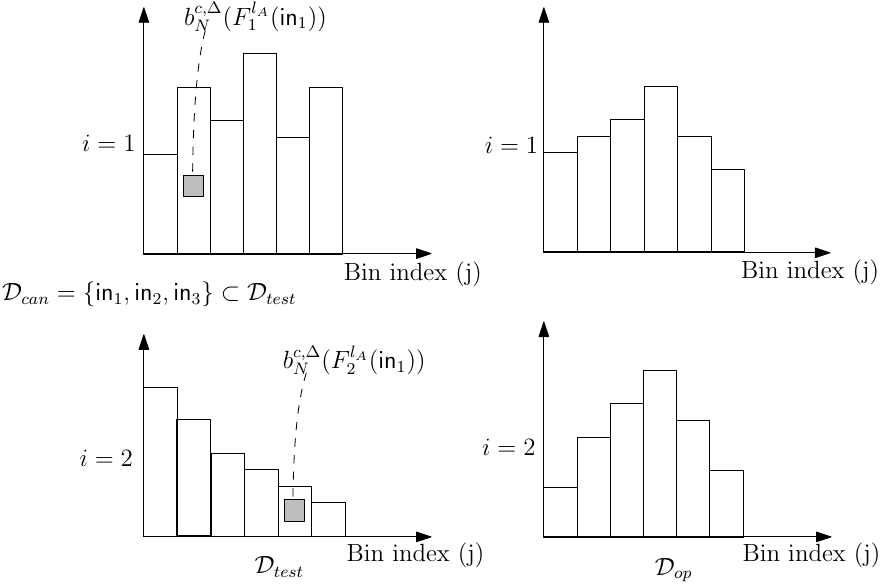}
    \caption{Shaping the test set by removing or adding data points requires simultaneously considering the effect of multiple neurons. For example, although removing $\sig{in}_1$ can reduce the count in bin $j=2$ for $i=1$, it may also undesirably reduce the count in bin $j=4$ for $i=2$.  The counts in the Y-axis are not on the same scale but only show the tendency.}
    \label{fig:multi.box}
    \vspace{-3mm}
\end{figure}

\vspace{3mm}

\section{MILP Encoding for Distribution Reshaping}
\label{sec:encoding}

Provided that $\mathcal{D}_{op}$ and $\mathcal{D}_{test}$ are not similar in distribution, we are interested in finding a subset of $\mathcal{D}_{test}$, such that the subset has the similar distribution with $\mathcal{D}_{op}$. However, this is not an easy task, as demonstrated in the example in Figure~\ref{fig:multi.box}: When one only looks at the distribution in~$i=1$ and removes data point~$\sig{in}_1$ in $\mathcal{D}_{test}$ to create distribution similarity, it can create a negative impact on~$i=2$, where the reduction may not be desired.

When one finds such a subset, due to the distribution similarity one can estimate the performance. However, keeping the subset as large as possible is also desired. This leads to Definition~\ref{def:distribution.shaping}, where we artificially introduce $\mathcal{D}_{can}$ to restrict the set of data points further as candidates to be removed. When $\mathcal{D}_{can}$ equals $\mathcal{D}_{test}$, any data point within  $\mathcal{D}_{test}$ can be removed. As demonstrated in later paragraphs, in our MILP encoding scheme, the number of 0-1 integer variables introduced in the MILP equals the size of~$\mathcal{D}_{can}$. 

\begin{definition}[Distribution Reshaping for $\epsilon$-portion similarity via Test Set Reduction]\label{def:distribution.shaping}
    Provided that $\mathcal{D}_{op}$ and $\mathcal{D}_{test}$ are not $\epsilon$-portion similar, given $\mathcal{D}_{can} \subseteq \mathcal{D}_{test}$, find $\mathcal{D}^{opt}_{can}\subseteq \mathcal{D}_{can}$ such that 
   \begin{enumerate}
        \item $\forall i \in \{1, \ldots, d_l\}$, $\mathcal{D}_{op}$ and $\mathcal{D}_{test} \setminus \mathcal{D}^{opt}_{can}$ are 
 $\epsilon$-portion similar.
       \item The size of $\mathcal{D}^{opt}_{can}$ is minimum, among any other multiset $\mathcal{D}^{'}_{can}$ that also ensures the first condition. 
   \end{enumerate}
\end{definition}

 If we take a data point $\sig{in} \in \mathcal{D}_{can}$, pass it to the DNN and extracts $F^{l}(\sig{in})$,  apply the binning function on each computed neuron value $F^{l}_i(\sig{in})$,  this  leads to a vector $v_{\sig{in}} \defeq (b^{c, \Delta}_{N}(F^{l}_1(\sig{in})), \ldots, \\ b^{c, \Delta}_{N}(F^{l}_{d_l}(\sig{in}))) \in \{0, \ldots, N\}^{d_l}$, where the value in each dimension $i \in \{1, \ldots, d_l\}$ contains the associated binning information for the $i$-th output. 
 
 For every $\sig{in} \in \mathcal{D}_{can}$, in our MILP encoding, create a 0-1 integer variable $br_{\sig{in}}$ that controls the decision of removing data point $\sig{in}$ from $\mathcal{D}_{can}$.

\vspace{-2mm}
\begin{itemize}
    \item If $br_{\sig{in}} = 1$, then remove the data point $\sig{in}$ from $\mathcal{D}_{test}$. 
    \item If $br_{\sig{in}} = 0$, then keep the data point
    $\sig{in}$.

\end{itemize}
\vspace{-2mm}

For neuron $i$, recall that the number of elements  originally in bin $j$ equals $\ct(j, B^{l_A}_i(F, \mathcal{D}_{test}))$. For  data point $\sig{in} \in \mathcal{D}_{can}$, if $F^{l_A}_{i}(\sig{in})$ falls into bin $j$, then if we remove the data point, the size of remaining elements in bin~$j$ can be characterized as
$\ct(j, B^{l_A}_i(F, \mathcal{D}_{test})) - br_{\sig{in}}$. 
Therefore, depending on the decision whether elements in $\mathcal{D}_{can}$ are removed or not, the number of elements for neuron~$i$, bin~$j$ can be encoded as $\ct (j, B^{l_A}_i(F, \mathcal{D}_{test})) - \sum_{\sig{in} \in \mathcal{D}_{can} \;\text{s.t.}\; b^{c,\Delta}_N(F^l_i(\sig{in})) = j} br_{\sig{in}}$.

\begin{lemma}[MILP encoding] The problem in Definition~\ref{def:distribution.shaping} can be reduced to the following MILP problem.
\vspace{-2mm}
\begin{multline}\label{eq:milp}
\text{minimize} \sum_{\sig{in} \in \mathcal{D}_{can}} br_{\sig{in}} \;\;\text{s.t.} \\
\forall i\in \{1, \ldots, d_l\}, \forall j\in \{0, \ldots, N\}:  \\
-\epsilon \leq \frac{\emph{\ct}(j, B^{l_A}_i(F, \mathcal{D}_{op}))}{|\mathcal{D}_{op}|} - \frac{\emph{\ct} (j, B^{l_A}_i(F, \mathcal{D}_{test})) - \sum_{\sig{in} \in \mathcal{D}_{can} \;\text{s.t.}\; b^{c,\Delta}_N(F^l_i(\sig{in})) = j} br_{\sig{in}}}{|\mathcal{D}_{test}| - \sum_{\sig{in} \in \mathcal{D}_{can}} br_{\sig{in}}} \leq \epsilon
\end{multline}

\end{lemma}

\proof The encoding is straightforward, where the only difference with Equation~\ref{def:portion.similar} is (1) the update of the denominator from $|\mathcal{D}_{test}|$ to $|\mathcal{D}_{test}| - \sum_{\sig{in} \in \mathcal{D}_{can}} br_{\sig{in}}$, reflecting the potential decrease in the data points, and (2) the update of the nominator by subtracting $\sum_{\sig{in} \in \mathcal{D}_{can} \;\text{s.t.}\; b^{c,\Delta}_N(f^l_i(\sig{in})) = j} br_{\sig{in}}$, counting the potential decrease of the number of items in bin~$j$. The remaining task is to ensure that the encoding does not lead to non-linear constraints. This holds, as $\frac{\ct(j, B^{l_A}_i(F, \mathcal{D}_{op}))}{|\mathcal{D}_{op}|}$ is a constant, one can rewrite the inequality by multiplying $|\mathcal{D}_{test}| - \sum_{\sig{in} \in \mathcal{D}_{can}} br_{\sig{in}}$.  \qed

\begin{example}

Consider $\mathcal{D}_{can} \defeq \langle \sig{in}_1, \sig{in}_2, \sig{in}_3 \rangle $, where for each element in $\mathcal{D}_{can}$, its binning information is listed in Table~\ref{table:example.encoding}. Then consider $i=3$ and $j=4$ in Equation~\ref{eq:milp}, the inequality part can be rewritten into Equation~\ref{eq:example}, where $\frac{\ct(4, B^{l_A}_3(F, \mathcal{D}_{op}))}{|\mathcal{D}_{op}|}$, $|\mathcal{D}_{test}|$, and $\ct (4, B^{l_A}_3(F, \mathcal{D}_{test}))$ are constants that can be computed before the MILP encoding. As $\epsilon$ is also a constant, Equation~\ref{eq:example} can be rewritten into two linear constraints. 
\begin{equation}\label{eq:example}
\;\;\;\;\;\;\;\;   -\epsilon \leq \frac{\ct(4, B^{l_A}_3(F, \mathcal{D}_{op}))}{|\mathcal{D}_{op}|} - \frac{\ct (4, B^{l_A}_3(F, \mathcal{D}_{test})) - br_{\sig{in}_1} - br_{\sig{in}_3}}{|\mathcal{D}_{test}| - br_{\sig{in}_1} - br_{\sig{in}_2} - br_{\sig{in}_3}} \leq \epsilon
\end{equation}

\end{example}

\begin{remark} Whenever distribution reshaping does not involve multiple neurons, finding a subset of $\mathcal{D}_{test}$ to ensure $\epsilon$-portion similarity can be done efficiently (with a greedy algorithm),  and therefore no MILP encoding required.
Reconsider the example in Figure~\ref{fig:multi.box} where the goal is only to perform reshaping on $i=1$. One can simply remove $\sig{in}_1$, as there is no side effect that should be considered.   
\end{remark}

\section{Evaluation}
\label{sec:evaluation}

\begin{figure}[t]
\centering
\begin{subfigure}{.5\textwidth}
  \centering
  \includegraphics[width=\linewidth]{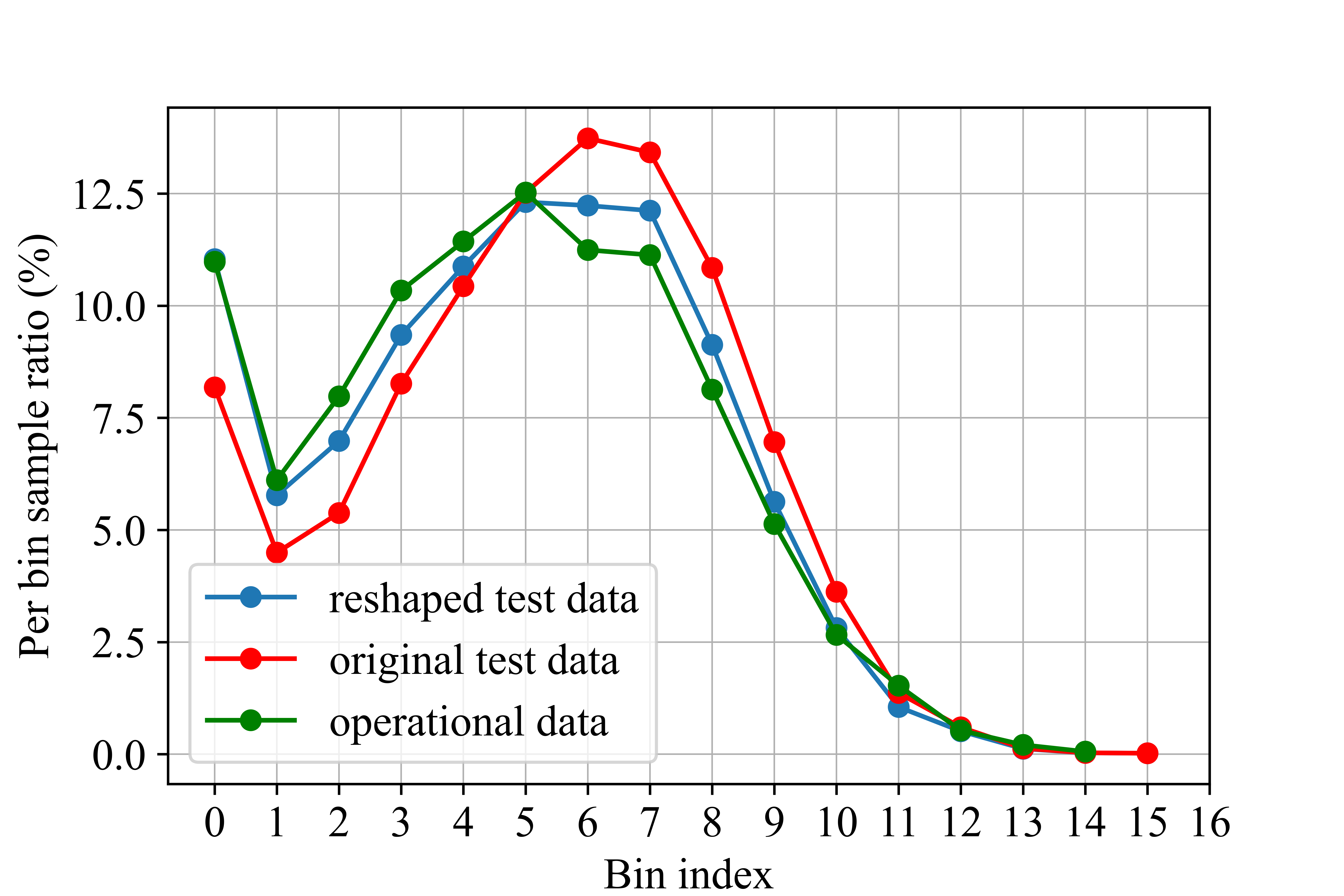}
  \caption{Activation values binned for  the \\ 5th neuron}
  \label{fig:result.output.reshaping.single}
\end{subfigure}%
\begin{subfigure}{.5\textwidth}
  \centering
  \includegraphics[width=\linewidth]{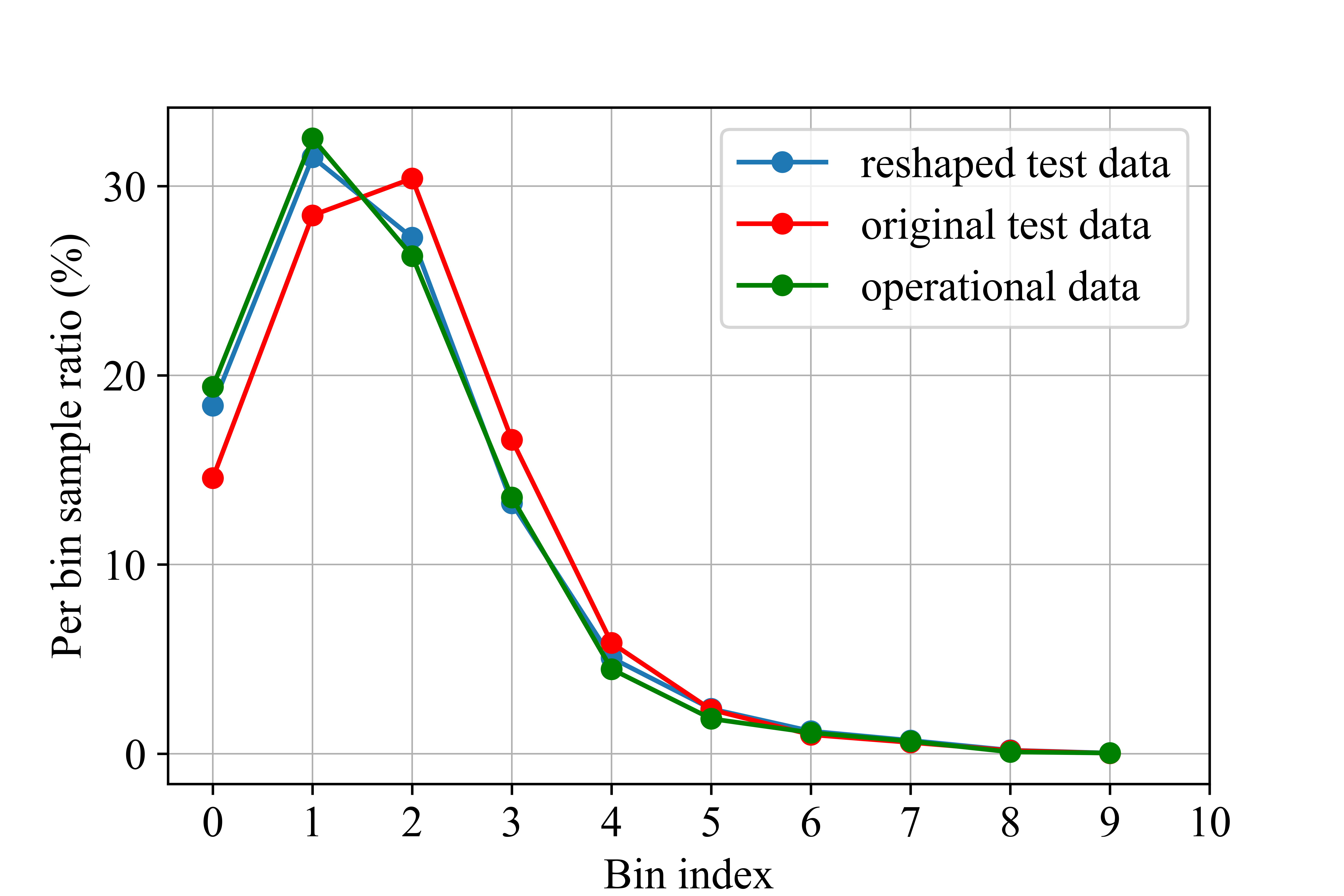}
  \caption{Activation values binned for the \\ 7th neuron}
  \label{fig:result.aggregated}
\end{subfigure}
\caption{Qualitative result of distribution reshaping simultaneously on multiple neurons}
\label{fig:effectiveness.of.distribution.reshaping}
\end{figure}

We have implemented the concept and performed an initial feasibility study based on the MNIST dataset~\cite{lecun1998mnist}\@. We use Pytorch~\cite{paszke2019pytorch} to train the DNN and use Google OR-Tools\footnote{\url{https://developers.google.com/optimization}} to solve the generated MILP problem. For distribution reshaping, we take $20$ neurons in close-to-output layers. To simulate the ``covariate shift", we have intentionally created the testing dataset with significantly more examples in classes ``7'', ``8'' and ``9'' (with each around~$20\%$), and left classes ``1'' to ``5'' have a small portion. The created operational dataset consists of $5300$ image samples\footnote{The set of $5300$ images is a set of data that we intentionally manipulate to create covariance shift and treat it as data being collected in the field.}, where the portion of ``7'', ``8'' and ``9'' is significantly reduced, meaning that the assumption on the frequency for class distribution is incorrect. Recall that in our problem definition, one does not have access to the operational data and the associated ground truth labels. The experimental setting here allows us to estimate whether distribution reshaping on neurons (representing the feature space) positively correlates with the data distribution reflected by the associated label. For measuring similarity, we use the bin width $\Delta = 1$ and set~$\epsilon$ to be $0.01$. The set $\mathcal{D}_{can}$, i.e., elements that can be removed for distribution reshaping purposes, ranges from~$7000$ to~$20000$. This implies that in the corresponding MILP problem, we have a maximum of~$20000$ binary integer variables. The whole program and the MILP solver are operated on an Intel~i5-9300H laptop equipped with 32GB RAM. Altogether the time required to find the smallest set to be removed for distribution reshaping is commonly below~$15$ minutes. Figure~\ref{fig:effectiveness.of.distribution.reshaping} shows the distribution for two neurons being considered. 

\begin{table}[t]
\centering
   \caption{An example for the MILP encoding}
\begin{tabular}{|c|c|c|c|}
\hline
 &  $b^{c,\Delta}_N(F^{l_A}_1(\cdot))$ & $b^{c,\Delta}_N(F^{l_A}_2(\cdot))$ & $b^{c,\Delta}_N(F^{l_A}_3(\cdot))$  \\ \hline
$\sig{in}_1$ & 1 & 4 & 4   \\ \hline
$\sig{in}_2$ & 2 & 1 & 2   \\ \hline
$\sig{in}_3$ & 0 & 3 & 4   \\ \hline
\end{tabular}
\label{table:example.encoding}
\end{table}

\begin{figure}[t]
\centering
\begin{subfigure}{.56\textwidth}
  \centering
  \includegraphics[width=\linewidth]{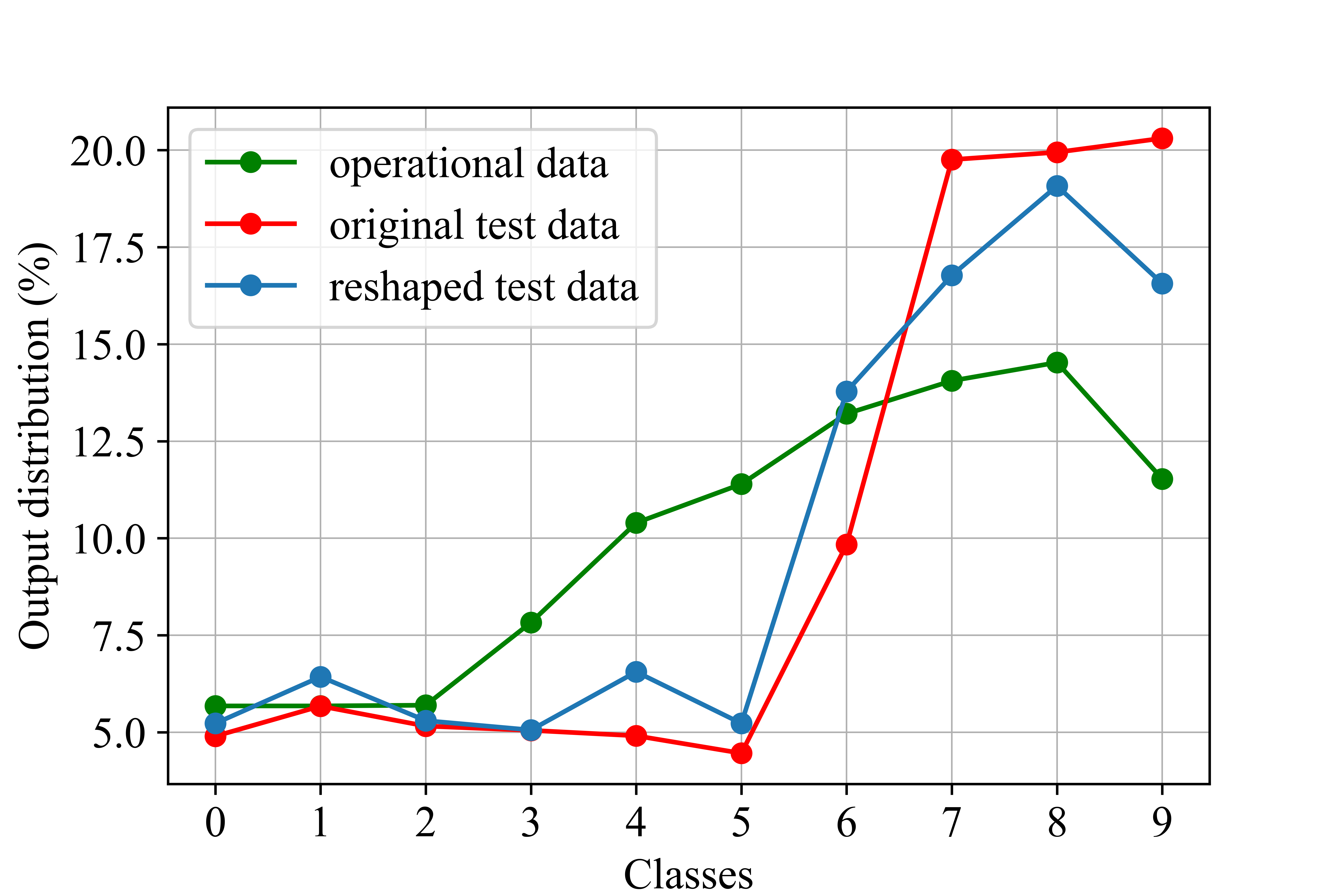}
  \caption{Output distribution in one experiment}
  \label{fig:single.output}
\end{subfigure}%
\begin{subfigure}{.4\textwidth}
  \centering
  \includegraphics[width=\linewidth]{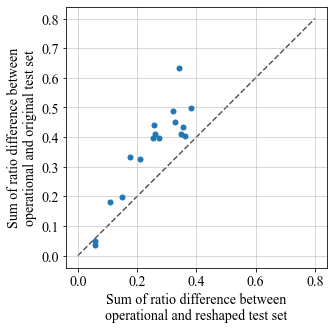}
  \caption{Aggregated result}
  \label{fig:aggregated.output}
\end{subfigure}
\caption{Effect of distribution reshaping, observed on the output label distribution}
\label{fig:distribution.reshaping.neurons}
\end{figure}

Figure~\ref{fig:distribution.reshaping.neurons}  presents our preliminary result, where in Figure~\ref{fig:single.output}, one can observe that although we only perform distribution reshaping on neurons reflecting the feature level, the reshaped test data is moving closer to the operational data when we consider the distribution reflected as the relative frequencies of each class, suggesting that the SPI estimation on the reshaped test dataset can be more precise. Figure~\ref{fig:aggregated.output} shows an aggregated result on all experiments being conducted, where the $x$ coordinate characterizes the sum of the per-class ratio-difference between operational and reshaped test data, and the $y$ coordinate characterizes the sum of the per-class ratio-difference between operational and the original test data. It also turned out that most of the points are within the top-left region, hinting that the reshaped test data demonstrates a positive correlation with the label distribution of the operational data.

\section{Concluding Remarks}
\label{sec:conclusion}

We investigated the problem of estimating the safety performance in the presence of covariate shift
based on test data and (feature-level) neuron value distributions -- but not on operational data, 
which often is unavailable in real-world situations.
Our main contribution is a MILP encoding for reshaping the test data to be similar to the (unknown) distribution 
of the operational data.
This reshaped test data now serves as a proxy for evaluating the safety performance 
in the presence of covariate shift. 
With this approach, we may  compute the distribution profile (as histograms) locally on the 
device of the DNN under investigation, thereby addressing possible privacy concerns. 
Initial experiments demonstrate the feasibility of this constraint-solving approach, but, 
clearly, more experience for more complex scenarios in real-world situations is needed.
However, the maximum number of removable samples is restricted, as this number correlates with
the  number of 0-1 variables in the generated MILP problem. 
Constraint solving, in particular, needs to be accelerated considerably.
Since the generated MILP encodings are highly-stylized, specialized variable branching heuristics
or suitable polynomial-time approximation schemes may be developed. 
Finally, the problem of reshaping test data can be generalized to also adapt individual test data points 
to covariate shift based on their respective resilience bounds~\cite{cheng2017maximum}\@.

\subsubsection*{Acknowledgements} This work is supported by the StMWi Bayern as part of the project for the thematic development of the Fraunhofer IKS.

 \bibliographystyle{plain}

\end{document}